\documentclass[letterpaper, 10 pt, conference]{ieeeconf}  % Comment this line out if you need a4paper

\usepackage{amsmath}
\usepackage{algorithm}
\usepackage{algpseudocode}
\usepackage{amsfonts}
\usepackage{graphicx}
\usepackage{cite} 
\usepackage{xcolor}
\usepackage{pifont}
\usepackage{booktabs}

\newcommand{\cmark}{\textcolor{green!60!black}{\ding{51}}}
\newcommand{\xmark}{\textcolor{red}{\ding{55}}}
\newcommand{\bx}{\mathbf{x}}

\newcommand{\bu}{\mathbf{u}}
\newcommand{\bff}{\mathbf{f}}

\newcommand{\btau}{\boldsymbol{\tau}}
\newcommand{\bxi}{\boldsymbol{\xi}}
\newcommand{\bnu}{\boldsymbol{\nu}}

\newcommand{\bmu}{\boldsymbol{\mu}}
\newcommand{\bpi}{{\boldsymbol \pi}}

\newcommand{\bbeta}{{\boldsymbol \beta}}

\newcommand{\matr}[1]{\mathbf{#1}} 

\newcommand{\bSigma}{\boldsymbol{\Sigma}}

\newcommand{\vect}[1]{\mathbf{#1}}

\DeclareMathOperator*{\argmin}{arg\,min}

\IEEEoverridecommandlockouts                              % This command is only needed if 
\title{\LARGE \bf Stochastic Multiple Shooting Trajectory Optimization via Sequential Local Policy Evaluation}

\author{Ashwin Gupta and Joseph Moore
\thanks{
Johns Hopkins University Whiting School of Engineering \newline \hspace*{1.6em} {\tt\small \{agupt139,jlmoore@\}@jh.edu}
\newline \hspace*{0.8em}
}
}
\renewcommand{\baselinestretch}{0.98}
\begin{document}
%\begin{titlepage}
%\vspace*{\fill}
%{\large
%\copyright 2025 IEEE.  Personal use of this material is permitted. Permission from IEEE must be obtained for all other uses, in any current or future media, including reprinting/republishing this material for advertising or promotional purposes, creating new collective works, for resale or redistribution to servers or lists, or reuse of any copyrighted component of this work in other works.}
%\vspace*{\fill}
%\end{titlepage}

\maketitle
\thispagestyle{empty}
\pagestyle{empty}

%%%%%%%%%%%%%%%%%%%%%%%%%%%%%%%%%%%%%%%%%%%%%%%%%%%%%%%%%%%%%%%%%%%%%%%%%%%%%%%%
\begin{abstract}

Stochastic single shooting trajectory optimization methods such as Model Predictive Path Integral control (MPPI) have been widely adopted in robotics due to their ability to reason about probabilistic dynamics and provide solutions where model gradients are noisy, costly to evaluate, or unavailable. However, satisfaction of terminal constraints when shooting over long action sequences is often sample inefficient, requiring a large number of iterations for convergence. In this paper, we present a stochastic multiple shooting method that optimizes short control action sequences connected via local feedback policies to improve sample efficiency and convergence to a terminal set. Additionally, we show that we are able to synthesize approximate system Jacobians purely from rollouts, making the method suitable for model-based reinforcement learning with black-box dynamics. We demonstrate the algorithm has improved sample efficiency and terminal set convergence for three nonlinear, underactuated optimization problems: a classic cartpole swingup task with analytical dynamics, a cartpole swingup task with learned neural network dynamics, and a VTOL quadplane performing a high angle-of-attack, precision post-stall landing maneuver.
\end{abstract}

%%%%%%%%%%%%%%%%%%%%%%%%%%%%%%%%%%%%%%%%%%%%%%%%%%%%%%%%%%%%%%%%%%%%%%%%%%%%%%%%
\section{INTRODUCTION}

Trajectory optimization is a fundamental task for robot planning and control with a broad array of applications ranging from nonlinear model predictive control (NMPC) on aerial vehicles \cite{basecu2024swarm} to model-based reinforcement learning (RL) for legged locomotion \cite{yang2019dataefficientreinforcementlearning}. Traditionally, trajectory optimization techniques have leveraged gradient-based numerical optimization routines such as sequential quadratic programming (SQP) \cite{bock1984} or interior point methods \cite{ipopt}. However, more recently, stochastic, sampling-based optimization methods like Model Predictive Path Integral Control (MPPI) \cite{mppi} and the Cross Entropy Method (CEM) \cite{kobilarov2012cem} have risen to prominence aided by the parallel processing afforded by modern Graphics Processing Unit (GPU) architectures. These stochastic optimization techniques do not require differentiable dynamics and can reason about probabilistic dynamics models to produce trajectories that are robust in expectation. 

While stochastic optimization is an extremely powerful technique for robotics, it is not without limitations. Most stochastic methods are classified as single shooting methods, where a single forward simulation of the system from a sequence of control inputs and an initial state produces the state trajectory. Single shooting is known to be sensitive to the optimization parameters, since small perturbations early in the trajectory can lead to large changes in the costs and constraints \cite{betts1998survey}. This sensitivity can render constraints, particularly terminal constraints far along the trajectory, very difficult to satisfy. %Moreover, since dynamic feasibility between knots is implicitly enforced by forward simulation, the optimizer may struggle to satisfy other constraints in the problem, particularly terminal constraints far along the trajectory.

Multiple shooting methods help alleviate these sensitivity issues for gradient-based solvers. In multiple shooting, the algorithm breaks the trajectory into shorter segments, applies shooting to each of these segments to improve numerical conditioning, and introduces a defect constraint at each boundary to ensure continuity \cite{betts1998survey}. However, few approaches for multiple shooting exist in the stochastic optimization literature as there are few gradient-free mechanisms by which to enforce hard constraints at defects. The Trajectory Bundle Method \cite{tracy2025trajectorybundle} is one sampling-based approach capable of enforcing defect constraints via sequential convex programming (SCP) and system rollouts. However, to our knowledge, the Trajectory Bundle Method does not address optimization for stochastic dynamics with process noise.

\begin{figure}[!t]%
  \centering
    \includegraphics[width=1.0\columnwidth]{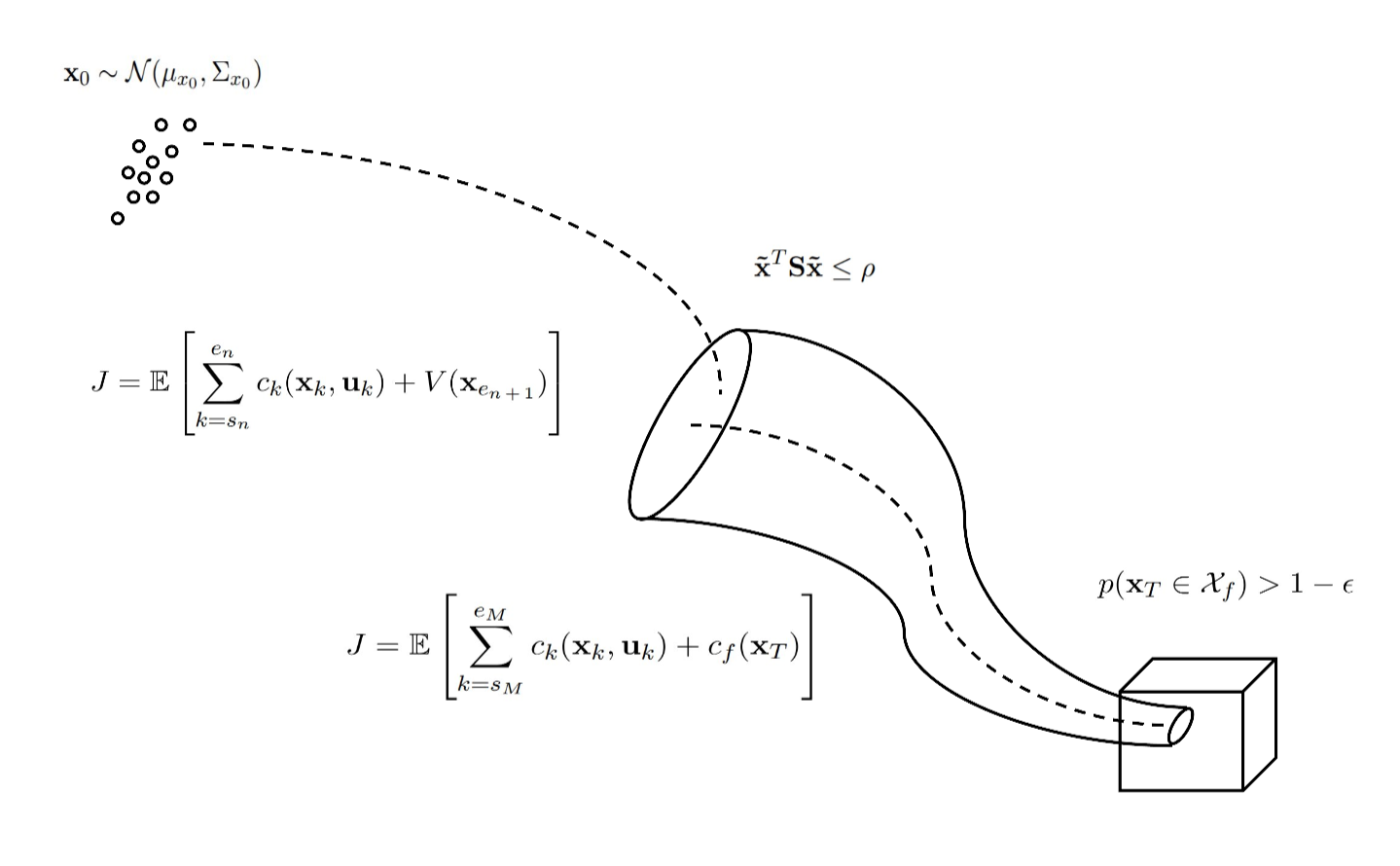}
    \caption{Illustration of the multiple shooting approach in state space on the terminal and penultimate segment. The terminal segment is optimized until convergence into the desired terminal set. A local time varying LQR feedback policy is built and the preceding segment and its initial condition are optimized to reach an ellipsoid level set of the cost-to-go such that the system will reach the terminal set under the feedback policy.}
  \label{fig:cover}
\end{figure}

In this paper, we introduce a multiple shooting framework for stochastic trajectory optimization. Our method maintains the beneficial properties of stochastic optimization, but employs multiple shooting to improve terminal set feasibility. Instead of formulating defect constraints as equality constraints, we formulate them as \emph{inequality constraints} by using a sequence of local time-varying feedback policies where the terminal state of a segment lies in the invariant set of the subsequent segment. Furthermore, we demonstrate how to construct these feedback policies purely from system rollouts so that no analytical model of the dynamics is required.

Our primary contributions are:
\begin{itemize}
\item A novel, gradient-free stochastic trajectory optimization method capable of solving high dimensional problems governed by a large class of nonlinear stochastic dynamics models.
\item Trajectory optimization results on two complex, underactuated systems demonstrating better terminal set convergence than baseline stochastic single shooting methods.
\item Trajectory optimization results on a non-differentiable neural network dynamics model with process noise.
\end{itemize}

\section{RELATED WORK}
Over the years, a wide variety of methods for numerical trajectory optimization have been proposed. Typically, these methods are separated into \emph{direct} and \emph{indirect} approaches. Researchers have also explored a variety of methods for trajectory optimization for stochastic dynamical systems.

\subsection{Direct Methods}
Direct trajectory optimization methods usually transcribe the original trajectory optimization problem directly as a nonlinear program \cite{betts1998survey}. Direct shooting methods parameterize the optimization problem using a sequence of control inputs and rely on forward simulation to satisfy the dynamics constraints \cite{kraft1985converting}. Direct multiple shooting methods divide the time horizon into segments, integrate over the smaller time intervals and incorporate state defect constraints \cite{bock1984multiple}. Oftentimes, the initial states for each segment are included as decision parameters. Direct collocation also include states and actions as decision parameters in the nonlinear program \cite{pardo2016evaluating, Bordalba_2023}. Dynamics constraints are enforced by a collocation scheme that can be transcribed into equality constraints that enforce derivative matching at particular points \cite{dircol}. These problems are then solvable by standard numerical optimization routines. Direct collocation methods allow for strict enforcement of terminal constraints as well as path constraints.

\subsection{Indirect Methods}
Indirect methods require the derivation of the necessary conditions for optimality, which are then satisfied numerically. A variety of indirect trajectory optimization approaches have been proposed over the years, including shooting, multiple shooting, and collocation approaches \cite{rao2010numericalmethods}. Differential dynamic programming (DDP) \cite{Mayne1966ASG} is an indirect shooting method that is common in robotics literature. Multiple shooting for DDP has been explored in various works \cite{Howell-2019-122091}\cite{li2023unifiedperspectivemultipleshooting} \cite{PELLEGRINI2020686} and has demonstrated improved convergence rate \cite{li2023unifiedperspectivemultipleshooting}. However, while in general indirect methods afford greater solution accuracy, they can be very sensitive to initialization and small changes in the boundary conditions \cite{bryson2018applied}.

%However, indirect methods require substantial analytical effort making it difficult to derive boundary value problems for nonlinear, stochastic systems.

\subsection{Stochastic Methods}
Researchers have explored a number of approaches for trajectory optimization for stochastic systems. Iterative LQG \cite{todorov2005generalized} and stochastic DDP \cite{todorov2005generalized} leverage local polynomial approximations of the stochastic dynamics. Other approaches such as \cite{ozaki2020tube, howell2021direct} assume local Gaussian stochastic dynamics and rely on the unscented transform to capture the propagation of uncertainty. Recently, with the advancement of GPUs, sampling-based approaches for stochastic trajectory optimization have become increasingly popular in the robotics community. 

Cross entropy trajectory optimization \cite{kobilarov2012cem} is one such approach that transforms the trajectory optimization problem into a rare-event probability estimation problem which can be solved heuristically via sampling. MPPI is another sampling-based trajectory optimization approach introduced as an optimal feedback controller for a particular class of control-affine dynamical systems in \cite{mppi}, and has since become ubiquitous in the robotics community. Many variants of MPPI have been proposed to improve robustness and generalize to different classes of systems \cite{wang2021variationalinferencempcusing}\cite{robustmppi}\cite{yin2022riskawaremodelpredictivepath}. More recent alternative stochastic trajectory optimization techniques have also been developed that demonstrate improved convergence, robustness \cite{polevoy2023probably}, and principled techniques for the selection of optimization parameters \cite{yi2024covompc}. Machine learning, particularly denoising diffusion \cite{pan2024modelbaseddiffusiontrajectoryoptimization}, has also shown promise for sampling-based trajectory optimization. The Trajectory Bundle Method \cite{tracy2025trajectorybundle} is a recent sampling-based approach capable of stochastic multiple shooting through the enforcement of defect constraints via SCP.

In this paper, we design a multiple shooting framework that is compatible with stochastic trajectory optimization algorithms. Fundamentally, our approach leverages a series of local feedback policies to satisfy defect constraints between segments and demonstrates improvements over single shooting baselines. We also propose an approach for generating these local policies using only samples.  

% \newline

\section{PROBLEM FORMULATION}
Our goal is to solve stochastic trajectory optimization problems of the form
%\begin{align}
%  & \min _ {\vect{u}_{0:T-1}} \mathbb{E} \left[ \sum_{k=0}^{T-1} c_k(\vect{x}_k, \vect{u}_k) + c_f(\vect{x}_{T}) \right] \\
%  \nonumber & s.t. \ \vect{x}_{k+1} = f(\vect{x}_k, \vect{u}_k, \vect{w}_k) \\  
%  \nonumber & \vect{x}_0 = \vect{x}_{\text{init}} \\
%  \nonumber & p(\vect{x}_T \in \mathcal{X}_f) > 1-\epsilon
%  \label{eq:optprob}
%\end{align}
\begin{equation}
\label{eq:optprob}
\begin{aligned}
\min_{\vect{u}_{0:T-1}} \quad
& \mathbb{E}\!\left[
    \sum_{k=0}^{T-1}
    c_k(\vect{x}_k,\vect{u}_k)
    + c_f(\vect{x}_T)
\right] \\
\text{s.t.}\quad
& \vect{x}_{k+1}
= f(\vect{x}_k,\vect{u}_k,\vect{w}_k), \\
& \vect{x}_0
= \vect{x}_{\text{init}}, \\
& p(\vect{x}_T \in \mathcal{X}_f)
> 1-\epsilon .
\end{aligned}
\end{equation}
where $f(\vect{x}_k, \vect{u}_k, \vect{w}_k)$ are the stochastic, discrete time dynamics of the system $\vect{x}_k \in \mathbb{R}^n, \vect{u}_k \in \mathbb{R}^m$, $c_k$ is the cost at time k, $c_f$ is the terminal cost, $\mathcal{X}_f$ is the desired terminal set, $\epsilon$ is a small positive constant, and $\vect{u}_{0:T-1}$ is the control input sequence.  It is assumed the terminal cost promotes convergence to the terminal set. For example, one may choose a level set of the terminal cost function.

Because we hypothesize that, like other multiple shooting methods, stochastic multiple shooting will lead to improved constraint satisfaction, we reformulate Eq. \ref{eq:optprob} as a general stochastic multiple shooting problem as follows:

Denote start times $s_1 ... s_M \in \mathbb{Z}$ and end times $e_1...e_M \in \mathbb {Z}$ where $M$ is the number of shooting segments. $e_{n}+1 = s_{n+1}$. Times are non-overlapping such that $0 = s_1 < e_1 < ... < s_M < e_M = T-1$.
%Let $n=\frac{T-1}{M}$ be a particular sub-interval length, where $M$ is the number of sub-intervals. We also define $N_j = nj$ and 
Let $\vect{X}_j = \vect{x}_{s_j}$, $\vect{U}_j=  \vect{u}_{s_j:e_j}$, and $\vect{W}_j=  \vect{w}_{s_j:e_j}$. We can now write:

%\begin{align}
%  \label{eq:multishoot}
%  \min _ {\vect{U}_{1:M},\vect{X}_{1:M}} &\mathbb{E} \left[ \sum_{j=1}^{M} \sum_{k=s_j}^{e_j} c_k(\vect{x}_k, \vect{u}_k) + c_f(\vect{x}_{T}) \right] \\\nonumber 
%  & \text{s.t.} \ \vect{X}_{j+1} = \mathcal{F}(\vect{X}_{j}, \vect{U}_j, \vect{W}_j) \\ \nonumber 
%  & \vect{X}_0 = \vect{x}_{\text{init}} \\\nonumber 
%  & p(\vect{x}_T \in \mathcal{X}_f) > 1-\epsilon.
%\end{align}
\begin{equation}
\label{eq:multishoot}
\begin{aligned}
\min_{\vect{U}_{1:M},\vect{X}_{1:M}} \quad
& \mathbb{E}\!\left[
    \sum_{j=1}^{M}
    \sum_{k=s_j}^{e_j}
    c_k(\vect{x}_k,\vect{u}_k)
    + c_f(\vect{x}_T)
\right] \\
\text{s.t.}\quad
& \vect{X}_{j+1}
= \mathbb{E}[\mathcal{F}(\vect{X}_j,\vect{U}_j,\vect{W}_j)], \\
& \vect{X}_0
= \vect{x}_{\text{init}}, \\
& p(\vect{x}_T \in \mathcal{X}_f)
> 1-\epsilon.
\end{aligned}
\end{equation}
where 
\begin{align*}
\mathcal{F}(\vect{X}_j,\vect{U}_j,\vect{W}_j)
={}&
f\!\Bigl(
    f\!\bigl(
        \cdots
        f(\vect{X}_j,\vect{u}_{s_j},\vect{w}_{s_j}),
        \ldots
    \bigr), \\
&\qquad
    \vect{u}_{e_j},
    \vect{w}_{e_j}
\Bigr).
\end{align*}

%\[
%f_{\vect{u},\vect{w}}(\vect{x})
%:=
%f(\vect{x},\vect{u},\vect{w}),
%\]
%\[
%\mathcal{F}(\vect{X}_j,\vect{U}_j,\vect{W}_j)
%=
%\left(
%f_{\vect{u}_{e_j},\vect{w}_{e_j}}
%\circ
%\cdots
%\circ
%f_{\vect{u}_{s_j},\vect{w}_{s_j}}
%\right)
%(\vect{X}_j).
%\]

$\vect{X}_{j+1} = \mathcal{F}(\vect{X}_{j}, \vect{U}_j, \vect{W}_j)$ is known as the defect constraint, and can be very difficult to satisfy without the formulation of hard constraints in the optimization problem \cite{tracy2025trajectorybundle}. %Moreover, for stochastic dynamics where $\vect{W}_j$ are random variables, exact defect satisfaction is often impossible even with the hard constraints in \cite{tracy2025trajectorybundle}.

%\begin{align}
%  &\min_{\vect{U}_{0:M-1},\vect{X}_{0:M-1}} \mathbb{E} \left[ \sum_{j=0}^{M-1}\sum_{k=N_j}^{N_{j+1}-1} c_k(\vect{x}_k, \boldsymbol{\pi}(\vect{x}_k)) + %c_f(\vect{x}_{T}) \right]\\\nonumber 
%  & \text{s.t.}\quad p(\mathcal{F}(\vect{X}_{j}, \vect{U}_j, \vect{W}_j)\in \mathcal{R}_{j+1})>1-\epsilon\\\nonumber 
%  & \vect{X}_0 = \vect{x}_{\text{init}}\\\nonumber 
%  & p(\vect{x}_T \in \mathcal{X}_f) > 1-\epsilon
%\end{align}

%A naive attempt to exploit this property would be to run a stochastic single shooting method for a few iterations to get $\vect{u}_{0:T-1}$, 
%fix the initial portion of the control tape up to some intermediate state $\vect{u}_{0:I}$ where $I < T-1$, and then
%continue to optimize the remaining portion of the trajectory $\vect{u}_{I+1:T-1}$ from the fixed intermediate state $\vect{x}_I$. However, it is not guaranteed that the terminal set is
%forward reachable under the dynamics from this intermediate state. Moreover, we would be limited to only two segments with this strategy.
% image: naive approach shortcoming
%These limitations can be resolved if we sample perturbations to the initial states of intermediate segments, but this introduces the additional complexity of satisfying defect constraints.

\section{APPROACH}
To solve the stochastic multiple shooting problem presented in Eq. \ref{eq:multishoot}, our approach leverages the following observation that the defect equality constraint can be relaxed into an \emph{inequality} constraint by leveraging a sequence of local feedback policies. Consider a local trajectory tracking policy $\bar{\bu}_k = \boldsymbol{\pi}_{j}(\bar{\bx}_k)$, where $\bar{\bx}_k$ and $\bar{\bu}_k$ denote the forward simulation of the state and input under a local trajectory tracking policy given by
\begin{align}
\vect{\bar{x}}_{k+1} = f(\bar{\bx}_k, \bpi(\bar{\bx}_k), \vect{w}_k).     
\end{align} If this policy is characterized by a non-empty local invariant set $\mathcal{R}_{j}$, we can now replace the defect equality constraint in Eq. \ref{eq:multishoot} with the inequality chance constraint 
\begin{align}
\quad p(\mathcal{F}(\vect{X}_{j}, \vect{U}_j, \vect{W}_j)\in \mathcal{R}_{j+1})>1-\epsilon_R,
\label{eq:defect}
\end{align}
where $\epsilon_R$ is a very small positive number. While the costs associated with $\bar{\bx}_k$ and $\bar{\bu}_k$ will not be identical to those associated with $\bx_k$ and $\bu_k$, we hypothesize that if the defects are small, the costs will be similar in magnitude.

Using the relaxation in Eq. \ref{eq:defect}, we construct an algorithm that uses stochastic optimization to generate trajectory segments whose final states are contained (with high probability) in the invariant sets associated with their subsequent segments. Not only can defect inequality constraint satisfaction be verified probabilistically, but our approach can also achieve local policy synthesis directly from samples. 

\begin{figure*}[t]
\vspace{2mm}
    \centering
    \includegraphics[width=\textwidth]{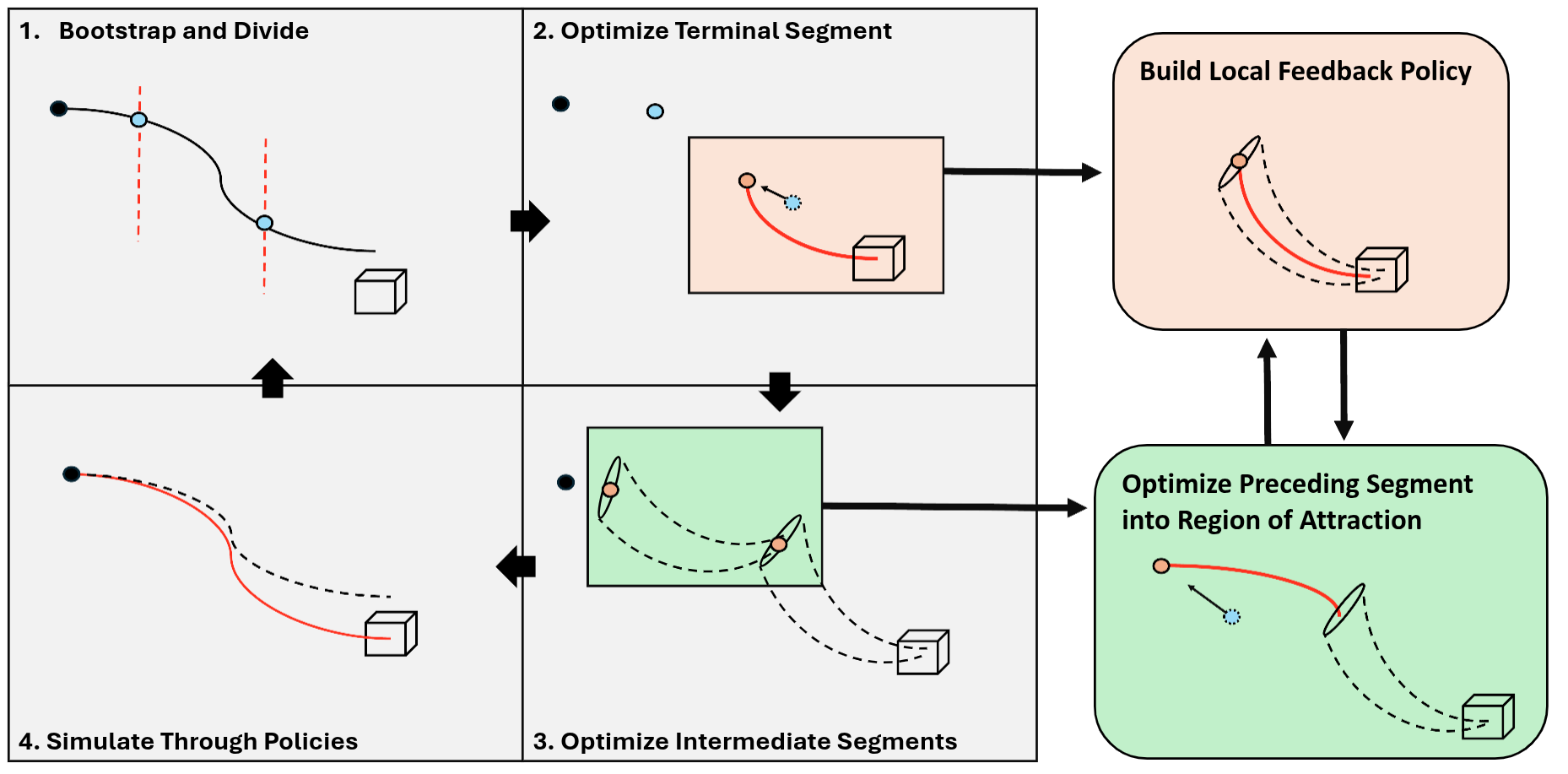}
    \caption{Overview of the multiple shooting framework. Blue dots indicate the intermediate states $\vect{x}_{s_n}$ from the bootstrapped trajectory $\vect{u}_{0:T-1}^-$(black line). The black dot is $\vect{x}_0$ and the black cube is $\mathcal{X}_f$. Optimized trajectories are indicated by red lines and their corresponding invariant sets under the feedback policy by dashed black lines. The steps in the green and orange boxes are repeated for each segment starting from the last and working backward until all segments have been optimized.
    }
    \label{fig:overview}
\end{figure*}
%\enlargethispage{\baselineskip}

An overview of the algorithm, illustrated in Fig. \ref{fig:overview}, is given as follows: 
\begin{enumerate}
% seed vs bootstrap
  \item Initialize the algorithm with an approximate candidate trajectory $\vect{u}_{0:T-1}^-$ and local feedback policy. This can be generated using a few iterations of stochastic single shooting or by warmstarting with a previous trajectory if operating in the context of NMPC.
  \item \label{step:divide} Divide the trajectory into segments with start times $s_1 ... s_M \in \mathbb{Z}$ and end times $e_1...e_M \in \mathbb {Z}$. $e_{n}+1 = s_{n+1}$ and segments are non-overlapping such that $0 = s_1 < e_1 < ... < s_M < e_M = T-1$.
  \item Beginning from the terminal segment, jointly optimize the initial condition $\vect{x}_{s_M}$ and the control inputs $\vect{u}_{s_M:e_M}$ until  convergence into the terminal set $\mathcal{X}_f$. Construct a local feedback policy.
  \item \label{step:recurse} Proceed backward to the preceding segment and jointly optimize the initial condition $\vect{x}_{s_{M-1}} $and control inputs $\vect{u}_{s_{M-1}:e_{M-1}}$ using the cost-to-go of the feedback policy for the subsequent segment as a terminal cost. Terminate when the invariant set of the subsequent feedback policy is reached. Construct a local feedback policy for the current segment.
  \item Repeat step \ref{step:recurse} until reaching the initial segment. For the initial segment, the initial condition $\vect{x}_0$ is held fixed.
  \item \label{step:last} Simulate the system forward from the initial state through the policies of each segment to produce the final complete control input sequence.
  \item Repeat steps \ref{step:divide} to \ref{step:last} $L$ times until convergence. This will be referred to as the ``outer loop".
  
\end{enumerate}

In the following subsections, we provide the additional details of our approach. Section \ref{sec:CEM} overviews our sampling-based trajectory optimization approach and Section \ref{sec:policy} presents our approach for constructing local feedback policies. Section \ref{sec:level-set-sample} provides a method for sampling initial segment states to promote inter-segment reachability, and section \ref{sec:traj-segments} discusses our approach for satisfying the defect inequality constraint. %Finally, \ref{sec:stochastic-verification} describes our approach 
While our algorithm uses the cross entropy method (CEM) to generate trajectory segments and the time-varying linear quadratic regulator (TVLQR) for the feedback policy, we note that our general method could be compatible with many stochastic optimization approaches and feedback policies, so long as the feedback policy has an associated cost-to-go.    
%A sampling strategy for intermediate states is presented in section \ref{sec:level-set-sample}. The defect constraint 
%is addressed in section \ref{sec:feedback-defect} by building a local feedback policy and optimizing the preceeding segment
%into the region-of-attraction of this policy. We verify the region of attraction as explained in \ref{sec:stochastic-verification} 
%to signal when to stop optimizing the current segment. Lastly, to eliminate dependence on analytical system Jacobians and permit for noisy, black-box
%dynamics models where finite differencing performs poorly, we present an approach for feedback policy synthesis purely from rollouts in section \ref{sec:least-sq}.

%For the remainder of this work, we will employ the cross entropy method for trajectory optimization, and the time-%varying linear quadratic regulator (TVLQR) for the feedback
%policy, but the generic skeleton of the approach is compatible with many choices of optimization algorithm and feedback policy provided one can construct or approximate a value function for said policy.

% TODO: diagram 
\subsection{Cross Entropy Trajectory Optimization}
\label{sec:CEM}
To optimize the trajectory segments, we employ Cross Entropy (CE) Trajectory Optimization \cite{kobilarov2012cem}. Because we modify the approach to optimize both a sequence of control inputs \emph{and} the segment's initial state to improve segment feasibility, we provide an review of the approach here. 

Consider the cost function $J(\btau,\bxi)$, where $\btau$ is a state trajectory given as $\btau=\{\bx_0, \bx_1 ...\allowbreak, \bx_{N_T} \}$ and $\bxi$ is an input sequence given as $\bxi=\{\bu_0, \bu_1 ...\allowbreak, \bu_{N_T-1} \}$. Let $\bxi$ be sampled from $\bxi\sim p(\bxi|\bnu)$, where $\bnu$ are the policy parameter distribution hyper-parameters. Let also $\bx_0$ be sampled from $\bx_0\sim p(\bx_0|\bbeta)$ where $\bbeta$ are the initial state distribution hyper-parameters. Trajectories are sampled from $\btau \sim p(\btau|\bxi,\bx_0)p(\bxi|\bnu)p(\bx_0|\bbeta)$, where
\begin{align}
p(\btau|\bxi,\bx_0)~=~p(\bx_0)\prod^T_{k=0}p(\bx_{k+1}|\bx_{k},\bu_k).
\label{eq:stochastic_dynamics}
\end{align} 
To minimize $\mathbb{E}_{\btau, \bxi, \bx_0 \sim p(\cdot, \cdot, \cdot|\bnu,\bbeta)}[J(\btau,\bxi)]$ and find the hyper-parameters $\bnu$ and $\bbeta$ we use CE. CE transforms the minimization of $J(\btau,\bxi)$ into a search for a the policy distribution $p(\bxi|\bnu)$ and initial condition distributions $p(\bx_0|\bbeta)$ by estimating the rare-event probability  
\begin{align}
\ell = \mathbb{P}_{\bar{\bnu},\bar{\bbeta}}(J(\btau,\bxi)\le \gamma) = \mathbb{E}_{\bar{\bnu},\bar{\bbeta}}[\mathbb{I}_{J(\btau,\bxi)\le\gamma}],   
\end{align} where $\gamma$ is a positive constant, $\mathbb{I}$ is the indicator function, and $\bar{\bnu}$ and $\bar{\bbeta}$ are a nominal sets of hyper-parameters. In this case, optimal solutions will correspond to events with very low probabilities, and the optimal parameters $\bnu^*$ and $\bbeta^*$ can be found by applying the Kullback-Leibler (KL) divergence as follows:%In particular, CE minimizes the Kullback-Leibler (KL) divergence between the optimal-importance sampling distribution $\frac{\mathbb{I}_{J(\btau)\le\gamma}p(\btau,\bar{\bnu})}{\ell}$ and $p(\bxi|\bnu)$, where $\bar{\bnu}$ is a nominal set of parameters. %given as 
%\begin{align}
%\bnu^* &= \arg\min_{\bnu} KL\bigg(\frac{\mathbb{I}_{J(\bZ)\le\gamma}p(\bz,\bar{\bnu})}%{\ell}, p(\bz,\bnu) \bigg ).
%\end{align}
%By applying the definition of the KL divergence, optimal parameters $\bnu^*$ are given as
\begin{align}
\hspace{-1mm}\bnu^*, \bbeta^*= \arg\max_{\bnu,\bbeta} \mathbb{E}_{\bar{\bnu}}\left[ \mathbb{I}_{J(\btau,\bxi) \leq \gamma} \log (p(\bxi|\bnu)p(\bx_0|\bbeta))\right].
\label{eq:importance}
\end{align}
%Because $p(\bxi|\bnu)$ and $p(\bx_0|\bbeta)$ are independent with separate parameters, this becomes 
%\begin{align}
%\bnu^*&= \arg\max_{\bnu} \mathbb{E}_{\bar{\bnu}}\left[ \mathbb{I}_{J(\btau,\bxi) \leq \gamma} \log %p(\bxi|\bnu)\right]\\
%\bbeta^*&= \arg\max_{\bbeta} \mathbb{E}_{\bar{\bbeta}}\left[ \mathbb{I}_{J(\btau,\bxi) \leq \gamma} \log %p(\bx_0|\bbeta)\right].
%\end{align}
Since that $p(\bxi|\bnu)$ and $p(\bx_0|\bbeta)$ are independent with separate parameters, Eq. \ref{eq:importance} can be approximated by sampling $N$ trajectories $\btau_i$ and $\bxi_i$ as follows:
\begin{align}
\hat{\bnu}^* = \arg\max_{\bnu} \frac{1}{N} \sum_{i=1}^N \mathbb{I}_{J(\btau_i,\bxi_i) \leq \gamma} \log p(\bxi| \bnu)\\
\hat{\bbeta}^* = \arg\max_{\bbeta} \frac{1}{N} \sum_{i=1}^N \mathbb{I}_{J(\btau_i,\bxi_i) \leq \gamma} \log p(\bx_0| \bbeta).
\end{align}
where $\bxi_i \sim p(\cdot|\bar{\bnu}) $ and $\bx_{0,i} \sim p(\cdot|\bar{\bbeta})$. Note that this formulation is equivalent to applying maximum likelihood estimation to a set of elite samples (i.e., $J(\btau_i,\bxi_i) \leq \gamma$). Because we cannot immediately sample from the optimal solution, CE applies an iterative approach to successively reduce the size of $\gamma$ until the parameter distributions converge. 

In this paper, we will parameterize the control sequence distribution $p(\bxi|\bnu)$ as a multivariate Gaussian $ \bxi \sim \mathcal{N}\left(\bxi | \bmu_u, \bSigma_u \right)$ with mean $\bmu_u$ and diagonal covariance $\bSigma_u$. We will parameterize the initial state distribution $p(\bx_0|\bbeta)$ as a multivariate Gaussian $ \bx_0 \sim \mathcal{N}\left(\bx_0 | \bmu_x, \bSigma_x \right)$ with mean $\bmu_x$ and covariance $\bSigma_x$.

%Consider the stochastic dynamics $p(\bx_{t+1}|\bx_{t}, \bu_{t})$ as defined in Section \ref{sec:prob_form}. Let us parameterize an open-loop control policy $\bu_t =\bpi(\btau) =\bzeta_t$, where $\bnu ~= ~\begin{bmatrix} \bzeta_0^T & \bzeta_1^T& ...&  \bzeta_{N_T}^T\end{bmatrix}^T$ and $\bzeta_t \in \mathbb{R}^{N_u}$. Let us also parameterize a distribution $p(\bxi|\bnu)$ as a multivariate Gaussian over the discrete-time control sequence so that $ \bxi \sim \mathcal{N}\left(\bU | \bmu, \bSigma \right)$. For computational tractability, we will also assume a diagonal covariance matrix $\bSigma$ so the distribution parameters are given as $\bnu \triangleq \begin{bmatrix} \bmu^T, diag(\bSigma)^T\end{bmatrix}^T$ where $ diag(\bSigma) = \begin{bmatrix} \bbeta_0^T & \bbeta_1^T& ...&  \bbeta_{N_T}^T\end{bmatrix}^T$ and $\bbeta_t \in \mathbb{R}^{N_u}$. We can then write a joint distribution
%$p(\bxi, \btau|\bnu)=p(\btau|\bxi)p(\bxi|\bnu)$, where $p(\btau|\bxi)$ is given by equation \ref{eq:stochastic_dynamics}.

\subsection{Local Feedback Policy}
\label{sec:policy}
To construct a local optimal feedback policy, we use discrete-time time-varying LQR (TVLQR). The policy is given as
\begin{align}
\bar{\bu}_k &= \matr{K}_k(\bar{\bx}_k-\bx_k)+\bu_k,
\end{align}
where the time-dependent feedback gain matrix $\matr{K}_k = (\matr{R}_c + \matr{B}_k^T\matr{S}_{k+1}\matr{B}_k)^{-1}(\matr{B}_k^T\matr{S}_{k+1}\matr{A}_k)$
and $\matr{S}_k$ is computed by integrating
\begin{align}
\matr{S}_{k-1} = &\matr{A}_k^T\matr{S}_k\matr{A}_k-...\nonumber\\&\matr{A}_k^T\matr{S}_k(\matr{R}_c+\matr{B}_k^T\matr{P}_k\matr{B}_k)^{-1}\matr{B}_k^T\matr{P}_k\matr{A}_k+\matr{Q_c}
\end{align}
backwards in time from $\matr{S}_N=\matr{Q}_{fc}\succeq0$. Here $\matr{A}_k=\frac{\partial \bff(\bx_k,\bu_k)}{\partial \bx_k}$ and $\matr{B}_k=\frac{\partial \bff(\bx_k,\bu_k)}{\partial \bx_k}$ and $\matr{Q}_c\succeq0$ and $\matr{R}_c\succ0$ are the weighting matrices on state and action respectively.
\subsection{Linear System Synthesis from Samples}
\label{sec:least-sq}
As seen above, TVLQR feedback policies require the Jacobians of the system $\matr{A}_k$ and $\matr{B}_k$, which may not always be available for non-smooth or noisy models such as neural network dynamics. Thus, in order to design a feedback policy, we employ a least-squares technique to synthesize a local linear system from random samples. 

Consider a nominal state
$\bx_k $, 
and nominal input 
$\bu_k$. We generate a set of $K$ nearby, perturbed initial states  $\mathcal{X} = \{\vect{x}^i\}_{i=1}^K \sim \mathcal{N}(\bx_k, \bSigma_x)$ 
and nearby, perturbed control input trajectories  $\mathcal{U} = \{\vect{u}^i\}_{i=1}^K \sim \mathcal{N}(\bu_k, \bSigma_u)$. We can write the linearized dynamics as
\begin{align}
    \dot{\vect{x}}_k \approx \vect{f}(\bx_k, \bu_k) 
    + \frac{\partial \vect{f}}{\partial \vect{x}_k} \, (\vect{x} - \bx_k) 
    + \frac{\partial \vect{f}}{\partial \vect{u}_k} \, (\vect{u} - \bu_k).
\end{align}
\\
Letting $\Delta \vect{x}^i = \vect{x}^i - \bx_k$ with $\vect{x}^i \in \mathcal{X}$  
and $\Delta \vect{u}^i = \vect{u}^i - \bu_k$ with $\vect{u}^i \in \mathcal{U}$ it follows

\begin{align*}
    &\begin{bmatrix}
    \dot{\vect{x}}^1 - \dot{\bx}_k & \dots & \dot{\vect{x}}^K - \dot{\bx}_k
    \end{bmatrix} \\
    &= \frac{\partial \vect{f}}{\partial \vect{x}_k} 
    \begin{bmatrix}
    \Delta \vect{x}^1 & \dots & \Delta \vect{x}^K
    \end{bmatrix}  +
    \frac{\partial \vect{f}}{\partial \vect{u}_k} 
    \begin{bmatrix}
    \Delta \vect{u}^1 & \dots & \Delta \vect{u}^K
    \end{bmatrix} 
    \\
    &= \begin{bmatrix} 
    \frac{\partial \vect{f}}{\partial \vect{x}_k} &
    \frac{\partial \vect{f}}{\partial \vect{u}_k}
    \end{bmatrix}
    \begin{bmatrix}
        \Delta \vect{x}^1 & \dots & \Delta \vect{x}^K \\
        \Delta \vect{u}^1 & \dots & \Delta \vect{u}^K
    \end{bmatrix}.
\end{align*}
From the above we can now construct the following least-squares problem to determine the system Jacobians:
\begin{align} \label{eq:jac}
    \argmin_{ 
    \frac{\partial \vect{f}}{\partial \vect{x}_k},
    \frac{\partial \vect{f}}{\partial \vect{u}_k} }
    \left\|
    \begin{bmatrix}
    (\dot{\vect{x}}^1 - \dot{\bx}_k)^T \\ \dots \\ (\dot{\vect{x}}^K - \dot{\bx}_k)^T
    \end{bmatrix}  - 
    \begin{bmatrix}
        \Delta \vect{x}^{1T} & \Delta \vect{u}^{1T} \\
        \dots & \dots \\
        \Delta \vect{x}^{KT} & \Delta \vect{u}^{KT}
    \end{bmatrix}
    \begin{bmatrix} 
    \frac{\partial \vect{f}}{\partial \vect{x}_k}^T \\
    \frac{\partial \vect{f}}{\partial \vect{u}_k}^T 
    \end{bmatrix}
    \right\|_2
    \nonumber
    \\
\end{align}

If the sparsity pattern of the Jacobians is known a priori, this can be enforced via equality constraints in the problem given by Eq. \ref{eq:jac}, and this problem can be solved with any standard convex optimization routine. We use CVXOPT \cite{andersen2013cvxopt} with the Splitting Cone Solver (SCS) \cite{scs}. 

\subsection{LQR Level Set Confidence Sampling} \label{sec:level-set-sample}
Since our stochastic optimization approach requires optimizing the initial state of a segment, it is important to provide a good initialization for both the mean and the covariance of the initial state distribution. A natural choice of distance metric for sampling perturbations to the initial state of the segment is one in which the 
system is contracting \cite{manchester2017controlcontractionmetricsconvex} under the current feedback policy. This lets us sample states that are ``close-by'' in the sense that the dynamics can be effectively steered by the inputs to minimize future cost. Since the cost-to-go for TVLQR is a valid contraction metric for the local linear system, we propose sampling states from a normal distribution with a P-th percentile confidence ellipse corresponding to some level set, $\rho$, of the cost-to-go.

We know that for a standard normal distribution the P-th quantile confidence ellipsoid is $ \vect{x}^T \vect{x} \leq {}^P\chi^2_n $, where $\vect{x} \in \mathbb{R}^n$ and ${}^P\chi^2_n$ is the P-th quantile of the chi-squared distribution with $n$ degrees of freedom. For an arbitrary zero mean normal distribution with covariance matrix $\matr{\Sigma}$, we use the z-score transformation $\vect{z} = \matr{\Sigma}^{-1/2} \vect{x}$ to transform the distribution to a standard normal distribution. Then the P-th quantile confidence ellipsoid is given by $\vect{x}^T \matr{\Sigma}^{-1} \vect{x} \leq {}^P\chi^2_n$. 

Thus, to generate a normal distribution whose \(P\)-th percentile confidence ellipsoid is given by $\vect{x}^T \matr{S} \vect{x} = \rho$ we can choose $\matr{\Sigma}^{-1}=\frac{{}^P\chi^2_n}{\rho}\,\matr{S}$.

%$\[ z^T z \leq {}^P\chi^2_n \implies x^T \Sigma^{-1} x \leq {}^P\chi^ 2_n \]
%Thus, to generate a normal distribution with the P'th percentile confidence ellipsoid to be given
%by $x^T S x = \rho$, we can choose $\Sigma ^{-1} = \frac{\chi^2_n}{\rho} S$, thus:
%\[ x^T \Sigma^{-1} x \leq {}^P\chi^2_n \implies x^T \frac{\chi^2_n}{\rho} S x \leq {}^P\chi^2_n \implies x^T S x = \rho \]

%\[
%\vect{z}^T \vect{z} \leq {}^P\chi^2_n
%\implies
%\vect{x}^T \matr{\Sigma}^{-1} \vect{x} \leq {}^P\chi^2_n
%\]

%thus

%\[
%\vect{x}^T \matr{\Sigma}^{-1} \vect{x}
%\leq
%{}^P\chi^2_n
%\implies
%\vect{x}^T
%\frac{{}^P\chi^2_n}{\rho}\,
%\matr{S}
%\vect{x}
%\leq
%{}^P\chi^2_n
%\implies
%\vect{x}^T \matr{S} \vect{x}
%\leq
%\rho.
%\]

\subsection{Optimizing Trajectory Segments} \label{sec:traj-segments}
Given a stochastic trajectory optimization approach and a strategy for feedback policy synthesis, we now discuss our approach for optimizing individual segments. Prior to optimizing any segments, we assume that a full-horizon local feedback policy warm-start has been generated via single shooting. For the terminal segment, we use the CEM to optimize both  $\bu_{s_M:e_M}$ and $\bx_{s_M}$ as follows:
\begin{align} \label{eq:terminal}
  & \min _ {u_{s_M:e_M}, x_{s_M}} \mathbb{E} \left[ \sum_{k=s_M}^{e_M} c_k(\vect{x}_k, \vect{u}_k) + c_f(\vect{x}_{T}) \right] \\
  \nonumber & s.t. \ \vect{x}_{k+1} = f(\vect{x}_k, \vect{u}_k, \vect{w}_k) \\ 
  \nonumber & p(\vect{x}_T \in \mathcal{X}_f) > 1-\epsilon .
\end{align}
In this case, the sampling for the distribution of $\bx_{s_f}$ is initialized by using the normal distribution constructed from the TVLQR cost-to-go level set provided by the warm-start policy as described in Section \ref{sec:level-set-sample}.   

To optimize non-terminal segments which span times $[s_n, e_n] \cap \mathbb{Z} $, we use the cost-to-go of the feedback policy for the succeeding segment as the terminal cost for the current segment. For the TVLQR policy discussed in Section \ref{sec:policy}, the cost-to-go at time $s_{n+1}$ is
\begin{align}
V(\vect{x}) = (\vect{x} - \vect{x}_{s_{n+1}})^T \matr{S} _{s_{n+1}} (\vect{x} - \vect{x}_{s_{n+1}}) 
= \tilde{\vect{x}}^T \matr{S}_{s_{n+1}} \tilde{\vect{x}}.
\end{align}
Thus, defect inequality constraints between segments can be enforced by solving the following optimization problem: 
\begin{align} \label{eq:intermediate}
  & \min _ {\vect{u}_{s_n:e_n}, \vect{x}_{s_n}} \mathbb{E} \left[ \sum_{k=s_n}^{e_n} c_k(\vect{x}_k, \vect{u}_k) + V(\vect{x}_{e_{n}\, + \, 1})  \right] \\ 
  \nonumber & s.t. \ \vect{x}_{k+1} = f(\vect{x}_k, \vect{u}_k, \vect{w}_k)\\
  \nonumber & p(\vect{x}_{e_n\, +\, 1} \in \mathcal{R}_{n+1}) > 1-\epsilon.
\end{align}
As in the case of the terminal segment, the sampling for the distribution of $\bx_{s_n}$ is initialized using the normal distribution constructed from the TVLQR cost-to-go level set provided by the warm-start policy described in Section \ref{sec:level-set-sample}.

\subsection{Probabilistic Constraint Satisfaction Criteria} \label{sec:stochastic-verification}

We consider constraint satisfaction achieved for the problems in Eq. \ref{eq:terminal} and Eq. \ref{eq:intermediate} when the system reaches the terminal set after being simulated forward using the subsequent sequence of local closed-loop policies. During CEM optimization of an individual segment, we construct a TVLQR policy about the current segment mean every $O$ iterations and rollout a batch of trajectories from the initial state of the current segment using current and subsequent feedback policies. If a sufficient percentage of these rollouts land in the terminal set, the segment iterations terminate. 

%Note that it may still be beneficial to iterate a few times beyond this condition to push the end of the current segment further into the region of attraction of the next segment's policy. 

In practice, we set a maximum CEM iteration limit and if the optimization does not converge within this limit, the algorithm executes further outer-loop iterations until the convergence condition is satisfied.

\subsection{Full Algorithm}
To construct our algorithm, we first define a few subfunctions. Let $\textsc{Rollout}(\vect{x}_0, \vect{u}_{0:T-1})$ perform an open loop rollout of the system from initial condition $\vect{x}_0$ and with control sequence $\vect{u}_{0:T-1}$. $\textsc{TVLQR}(\vect{x}_{0:T-1}, \vect{u}_{0:T-1}, \matr{Q}_c, \matr{R}_c, \matr{Q}_f)$ computes optimal linearized feedback gains $\matr{K}_{0:T-1}$ and cost-to-go matricies $\matr{S}_{0:T-1}$ from a nominal state trajectory and input trajectory as well as the gain matrices $\matr{Q}_c, \matr{R}_c, \matr{Q_f}$ as described in Sec. \ref{sec:policy}. $\textsc{CEM}(\vect{u}^-_{0:T-1}, \vect{x}^-_{0}, \matr{\Sigma}_{u},  \matr{\Sigma}_{x} )$ perfoms the cross entropy optimization described in Sec. \ref{sec:CEM} and returns the optimal control sequence $\vect{u}_{0:T-1}$ and initial state $\vect{x}_0$ given warm starts $\vect{u}^-_{0:T-1}$, and $\vect{x}^-_0$ and covariance matricies $\Sigma_{u}, \Sigma_{x}$. Finally, $\textsc{SimPolicy}(\vect{x}_0, \vect{x}_{0:T-1}, \vect{u}_{0:T-1}, \matr{K}_{0:T-1})$ performs a rollout from $\vect{x}_0$ using the given the nominal trajectory $\vect{x}_{0:T-1}, \vect{u}_{0:T-1}$ and feedback gains $\matr{K}_{0:T-1}$ of the closed loop policy as described in Sec. \ref{sec:policy}.
 The full algorithm is given in Alg. \ref{alg:ms}. This should be executed for $L$ outer loop iterations until convergence.

\begin{algorithm}[bth]
\caption{Stochastic Multiple Shooting}
\label{alg:ms}
\begin{algorithmic}[1]

\Require  $\vect{u}_{0:{T-1}}^-, \vect{x}_0, \Sigma_{u, 0:T-1}, \rho_{1:M}, \chi^2_n, \matr{Q}_c, \matr{R}_c, \matr{Q}_f$
\State $\vect{x}_{0:T}^- \gets \textsc{Rollout}(\vect{x}_0, \vect{u}_{0:{T-1}}^- )$
\State $\matr{\tilde{S}}_{0:T} \gets$ 
$\textsc{TVLQR}(\vect{x}^-_{0:T-1}, \vect{u}^-_{0:T-1}, \matr{Q}_c, \matr{R}_c, \matr{Q}_f)$
\State Choose $s_1, e_1, s_2, ... s_M, e_M$
%\State Solve Problem in Eq. \ref{eq:terminal} via Alg \ref{alg:cem_trajopt} 
%\[ 
%\vect{u}_{s_f:e_f} \gets \textsc{CEM}(\vect{u}^-_{s_f:e_f}, \vect{x}_0, \Sigma_{u,s_f:e_f},  \frac{\rho_f}{\chi^2_n} \vect{S}_{s_f}^{-1} ) 
%\]
%\State $\matr{K}_{s_f:e_f}$ $\matr{S}_{0:K-1}\gets$ 
%\Statex $\textsc{TVLQR}(\vect{x}^-_{0:K-1}, \vect{u}^-_{0:K-1}, \matr{Q}_c, \matr{R}_c, \matr{Q}_f)$
\For{$n = M,\dots,1$}
    \If{$n =M$}
    \State Solve Problem in Eq. \ref{eq:terminal}. %via Alg \ref{alg:cem_trajopt},
    \Else 
    \State Solve Problem in Eq. \ref{eq:intermediate}. %via Alg \ref{alg:cem_trajopt},
    \EndIf
    \State  
    $
    \vect{u}_{s_n:e_n}, \vect{x}_{s_f} \gets \textsc{CEM}(\vect{u}^-_{s_n:e_n}, \vect{x}_{s_n}^-, \Sigma_{u,s_n:e_n},  \frac{\rho_f}{\chi^2_n} \tilde{\vect{S}}^{-1}_{s_n} ) 
    $
    \State $\vect{x}_{s_n:e_{n}\, + \, 1} \gets \textsc{Rollout}(\vect{x}_{s_n}, \vect{u}_{s_n:e_n} )$
    \State $\matr{K}_{s_n:e_n}$ $\matr{S}_{s_n:e_n}\gets$ 
    \Statex $\textsc{TVLQR}(\vect{x}_{s_n:e_n}, \vect{u}_{s_n:e_n}, \matr{Q}_c, \matr{R}_c, \matr{Q}_f)$
\EndFor
\State $\vect{u}_{0:T-1}^+ \gets \textsc{SimPolicy}(\vect{x}_0, \vect{x}_{0:T-1}, \vect{u}_{0:T-1}, \matr{K}_{0:T-1})$
\State \Return $\vect{u}_{0:{T-1}}^+$

\end{algorithmic}
\end{algorithm}

\section{RESULTS}

\subsection{Cartpole}
We begin our evaluation with a classic cartpole dynamical system with state vector $\vect{x} = \begin{bmatrix} p & \theta & \dot{p} & \dot{\theta} \end{bmatrix}^T$ where $p$ is the cart position and $\theta$ is the pendulum angle. The swingup cost is given as 
\[
c_f(\vect{x}) = 100 p^2 + 1000 (\theta - \pi)^2 + 10  \dot{p}^2 + 10  \dot{\theta}^2 \\
\]
\[
c_k(\vect{x}, \vect{u}) = 0.01 u^2.
\]

We use an RK2 integrator with a time horizon of 35 knots with $\Delta t = 0.1$. The system has process noise with standard deviation 0.1 m/s/s and 0.05 rad/s/s. We use three segments of lengths 10, 10 and 15 knots for multiple shooting and 5 percent elites, and bootstrap multiple shooting with 5 iterations of single shooting. We use 4 outer loop iterations of Alg. \ref{alg:ms}. As a baseline, we use a standard single shooting cross entropy optimization with 5 percent elites, and MPPI tuned to the best temperature parameter we could find. 

For purpose of evaluation, we will define a ``rollout'' to be a single forward control trajectory simulation up to the \emph{full} time horizon $T$. Dynamics evaluations are parallelized over a batch dimension of states and controls, but not across timesteps. Therefore the total number of rollouts serves as a hardware and algorithm agnostic proxy for GPU compute time and permits for direct comparison of single and multiple shooting methods. 

For example, single shooting with a batch size of 100 and 30 CEM iterations would be 3000 rollouts. Similarly, for a trajectory with the same number of knots, multiple shooting with two even length segments, 3 outer loop iterations, and 10 CEM iterations per segment, would also be 3000 rollouts.

\begin{figure}[tbh]%
  \centering
    \includegraphics[clip, width=1.0\columnwidth]{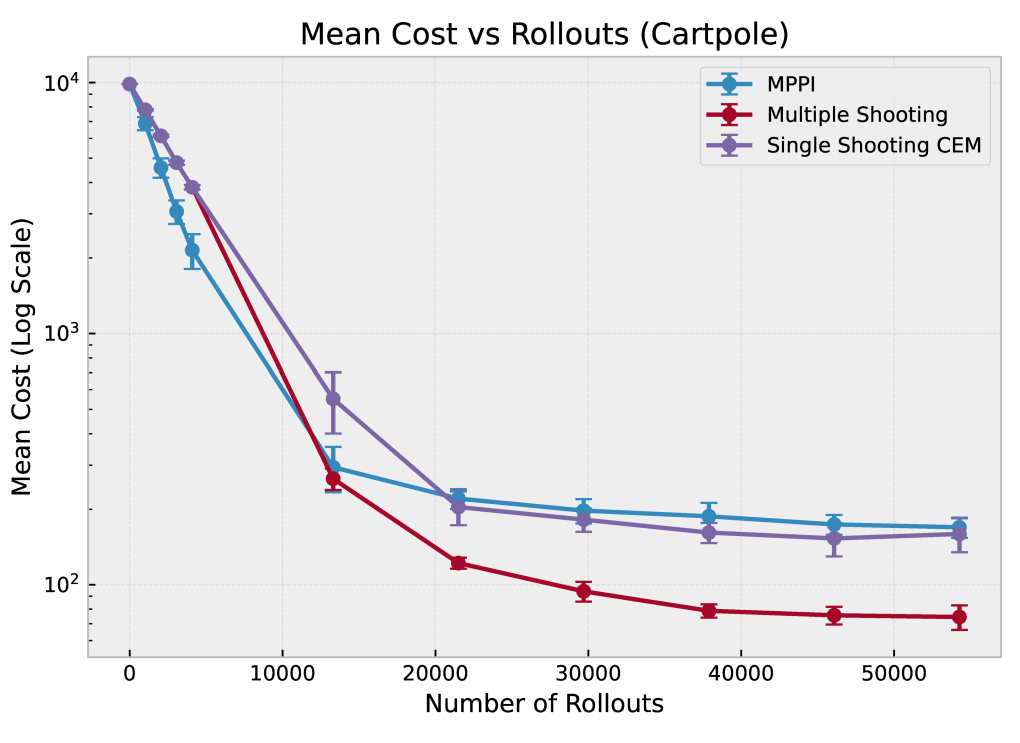}
    \caption{Number of rollouts vs mean cost for 10 trials of the cartpole swing-up problem using each optimization algorithm. Error bars indicate one standard deviation.}
  \label{fig:cartpole-iters}
\end{figure}

As shown in this example, the number of CEM optimizer iterations are not directly comparable across methods since multiple shooting optimizes over shorter trajectories. The mean cost versus number of rollouts for 10 trials of the optimization is shown in Fig. \ref{fig:cartpole-iters}.  The multiple shooting method converges faster than MPPI and single shooting CEM, and also achieves lower total cost at convergence.

The terminal set is defined to be a state space box with lower and upper bounds
$\vect{x}_L = \begin{bmatrix} -0.1, \pi - \frac{\pi}{12}, -0.3, -0.1\end{bmatrix}$ and 
$\vect{x}_U = \begin{bmatrix} 0.1, \pi + \frac{\pi}{12}, 0.3, 0.1\end{bmatrix}$ and only the multiple shooting method achieves convergence to the terminal box as seen in Table \ref{table:terminal}. The costs provided are the average terminal cost at convergence. The green check indicates convergence to the specified terminal box. A comparison of MPPI and Multiple shooting trajectories at convergence can be seen in Fig. \ref{fig:cartpole-iters}.

\begin{figure}[tbh]%
  \centering
    \includegraphics[clip, width=1.0\columnwidth]{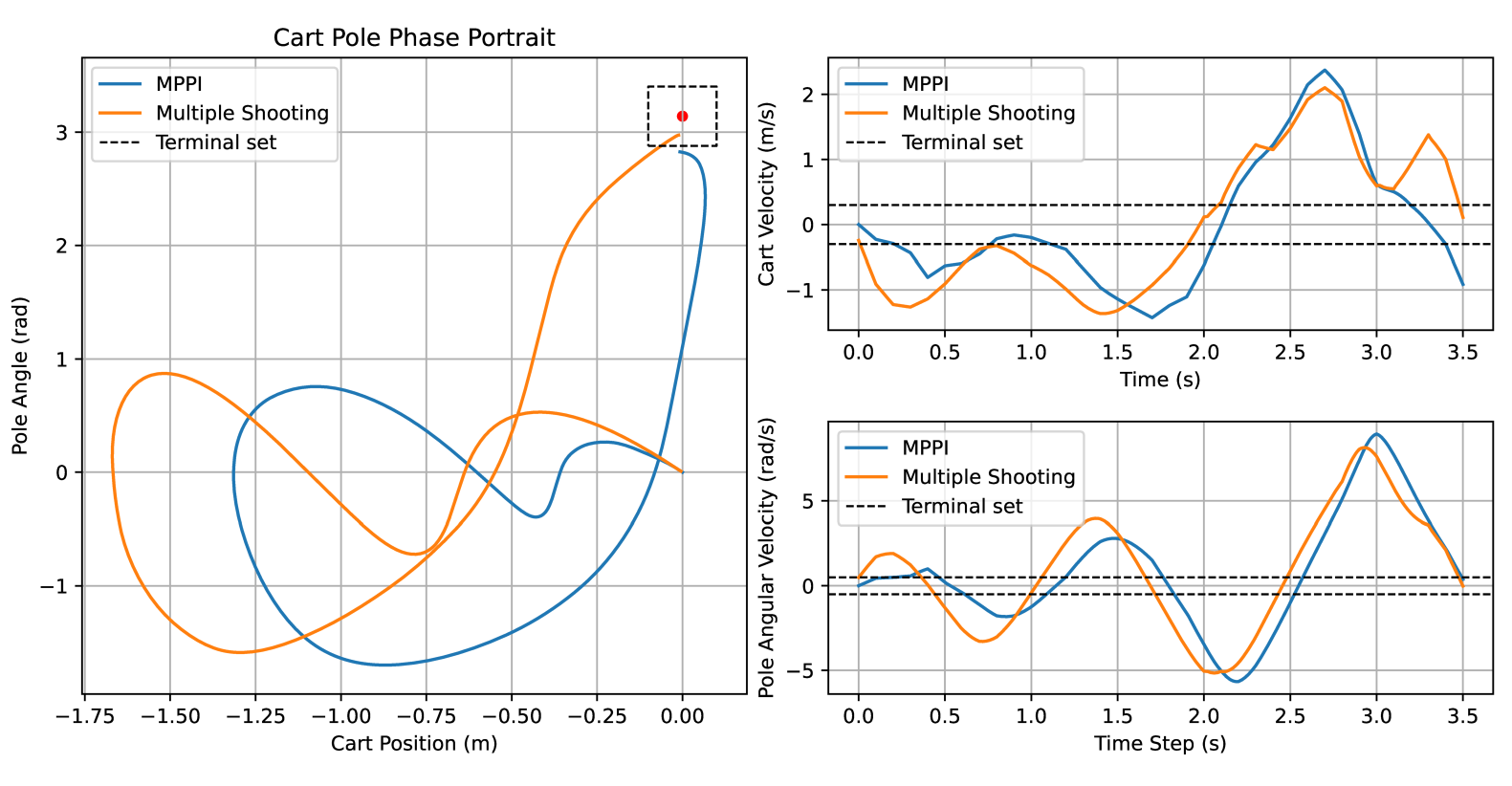}
    \caption{Cartpole Swing-Up Trajectories. The multiple shooting method enables convergence to the desired terminal set.}
  \label{fig:cartpole-traj}
\end{figure}

\subsection{Learned Cartpole}
\begin{figure}[tbh]%
  \centering
    \includegraphics[clip, width=1.0\columnwidth]{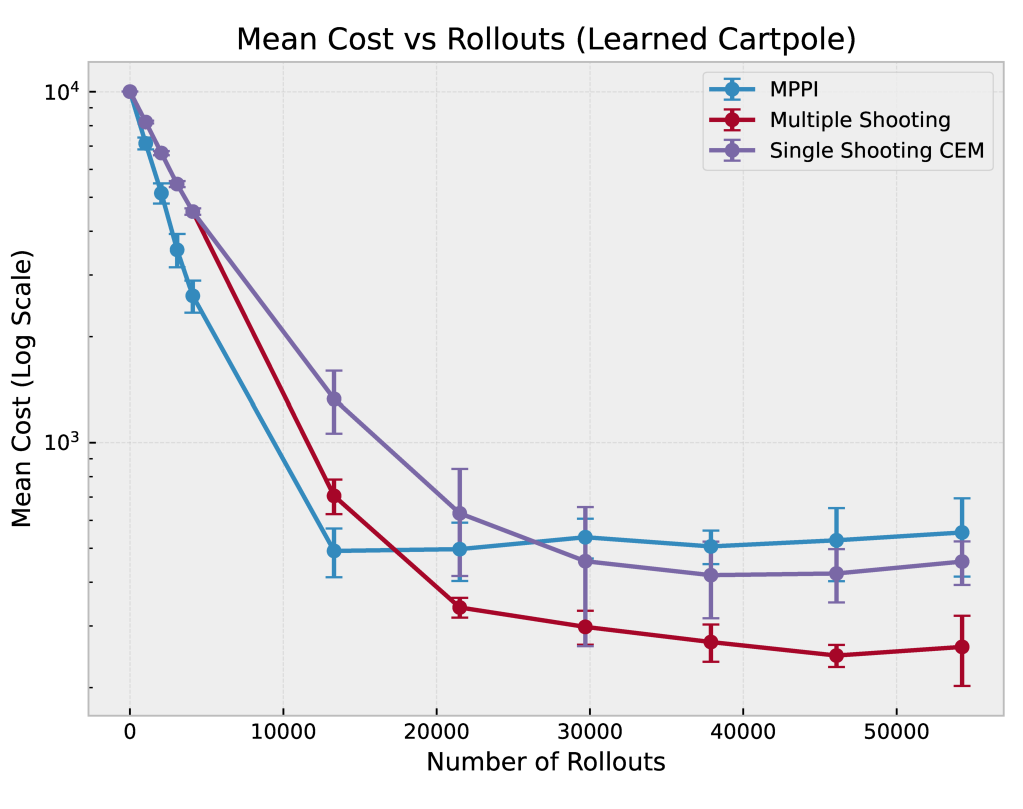}
    \caption{Number of rollouts vs mean cost for 10 trials of the cartpole swing-up problem on the learned model using each optimization algorithm. Error bars indicate one standard deviation.}
  \label{fig:learned-cartpole-iters}
\end{figure}

We consider a ``black-box'' model of the cartpole system. In this case we have a learned fully-connected neural network dynamics model with process noise. The learned model has two hidden layers of 256 and 512 neurons and uses ReLU activations. Additionally, we add zero-mean normally distributed process noise to the system at each step of the dynamics. The same optimization parameters and costs are applied to this system as the analytical cartpole model, and the results are shown in Fig. \ref{fig:learned-cartpole-iters}. We employ the least squares method for approximate Jacobians on this system since finite differencing through the noisy dynamics produces poorly conditioned Jacobians.

\subsection{VTOL Quadplane}
\begin{figure}[tbh]%
\vspace*{2mm}
  \centering
    \includegraphics[clip, width=1.0\columnwidth]{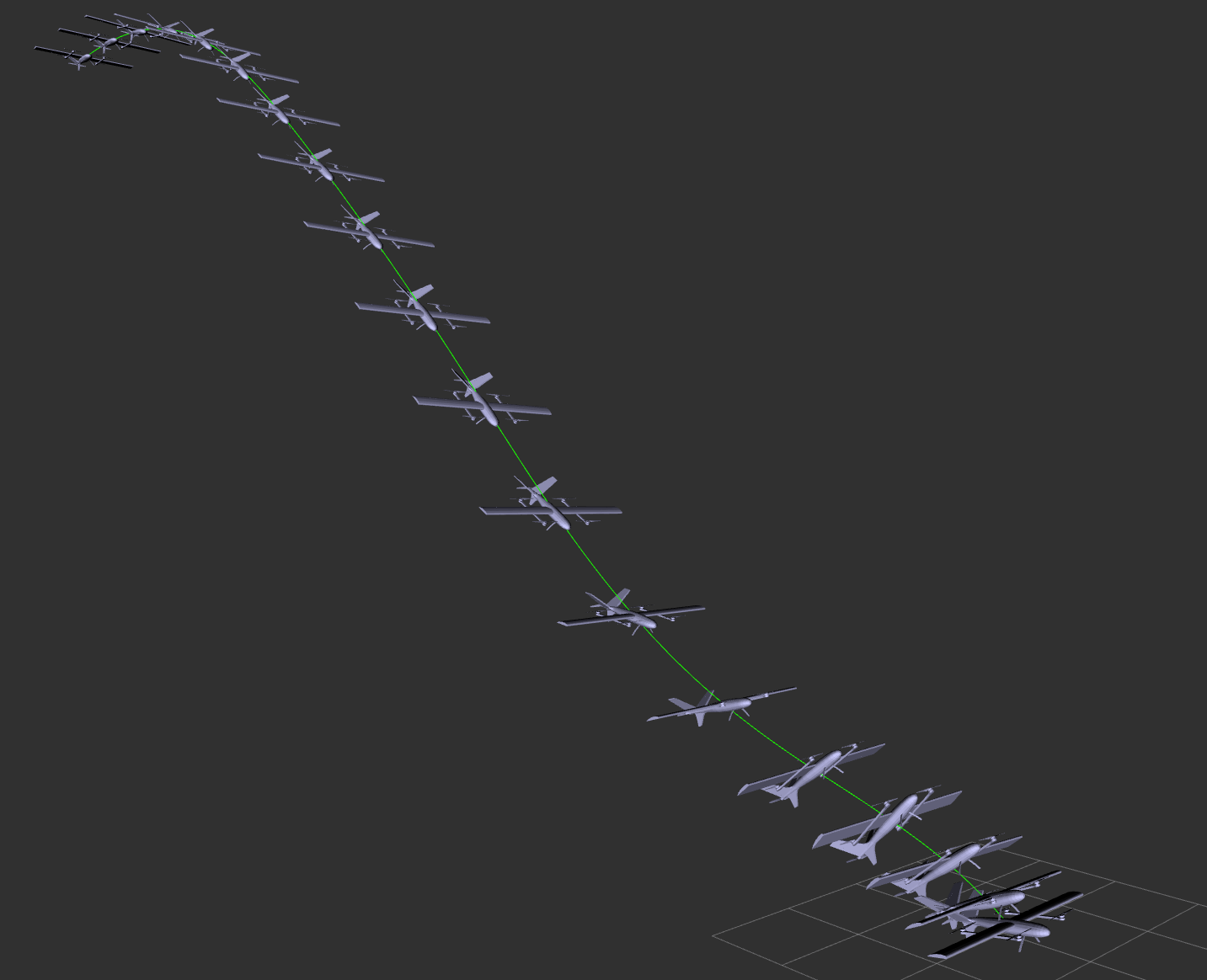}
    \caption{Visualization of quadplane precision landing trajectory. The system has a 22-dimensional state space and 9-dimensional input space. It lands 30 m downrange from 15 m altitude and 15 m/s initial cruise velocity in 2.5 seconds with sub 2 m/s velocity and within 1 m position error. }
  \label{fig:vtol-graphic}
\end{figure}

Finally, we evaluate the method on a 2 meter wingspan VTOL quadplane with a 22-dimensional state space and 9 dimensional input space corresponding to 5 propeller throttles, ailerons, elevator, and a rudder. The dynamics model is detailed in \cite{basecu2024swarm}, and this serves as a high dimensional nonlinear system with complex couplings between inputs and states. The goal is to perform a precision landing maneuver 30 meters downrange starting at 15 m/s level flight and 15 meter altitude. The aircraft maneuver is pictured in Fig. \ref{fig:vtol-graphic}.

The terminal set is 1.2 meter radius sphere for position, and the speed in any axis must not exceed 2.5 m/s. The total orientation error must be within 15 degrees of zero.  The terminal cost function is quadratic in position, velocity, and orientation and the running cost is quadratic in angular rate and control input. Additionally, we enforce collision-free trajectories with the ground using a very high binary cost. We use 2 segments for multiple shooting of 40 knots and 15 knots respectively at $\Delta t = 0.05$ with an RK2 integration scheme. The multiple shooting trajectory is initially bootstrapped by 10 iterations of single shooting. We use two outer loop iterations of Alg. \ref{alg:ms}. The system has process noise in acceleration and angular acceleration.

\begin{figure}[tbh]%
  \centering
    \includegraphics[clip, width=1.0\columnwidth]{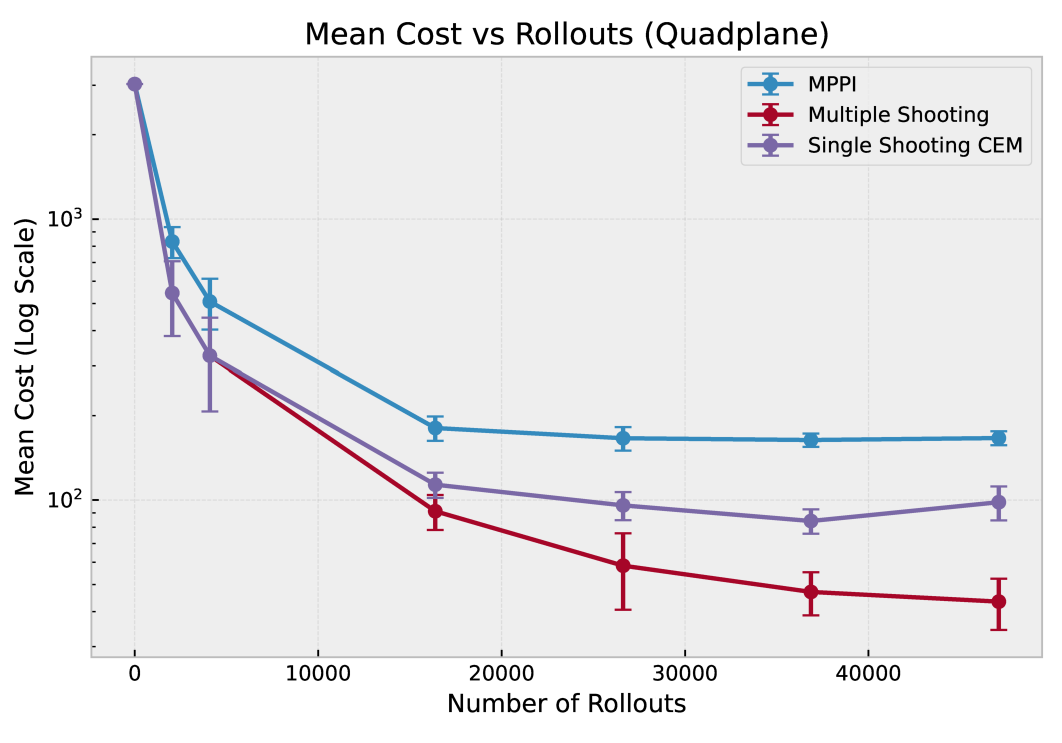}
    \caption{Number of rollouts vs mean cost for 10 trials of the VTOL landing problem using each optimization algorithm. Error bars indicate one standard deviation.}
  \label{fig:vtol-iters}
\end{figure}

\begin{figure}[tbh]%
\vspace*{2mm}
  \centering
    \includegraphics[clip, width=1.0\columnwidth]{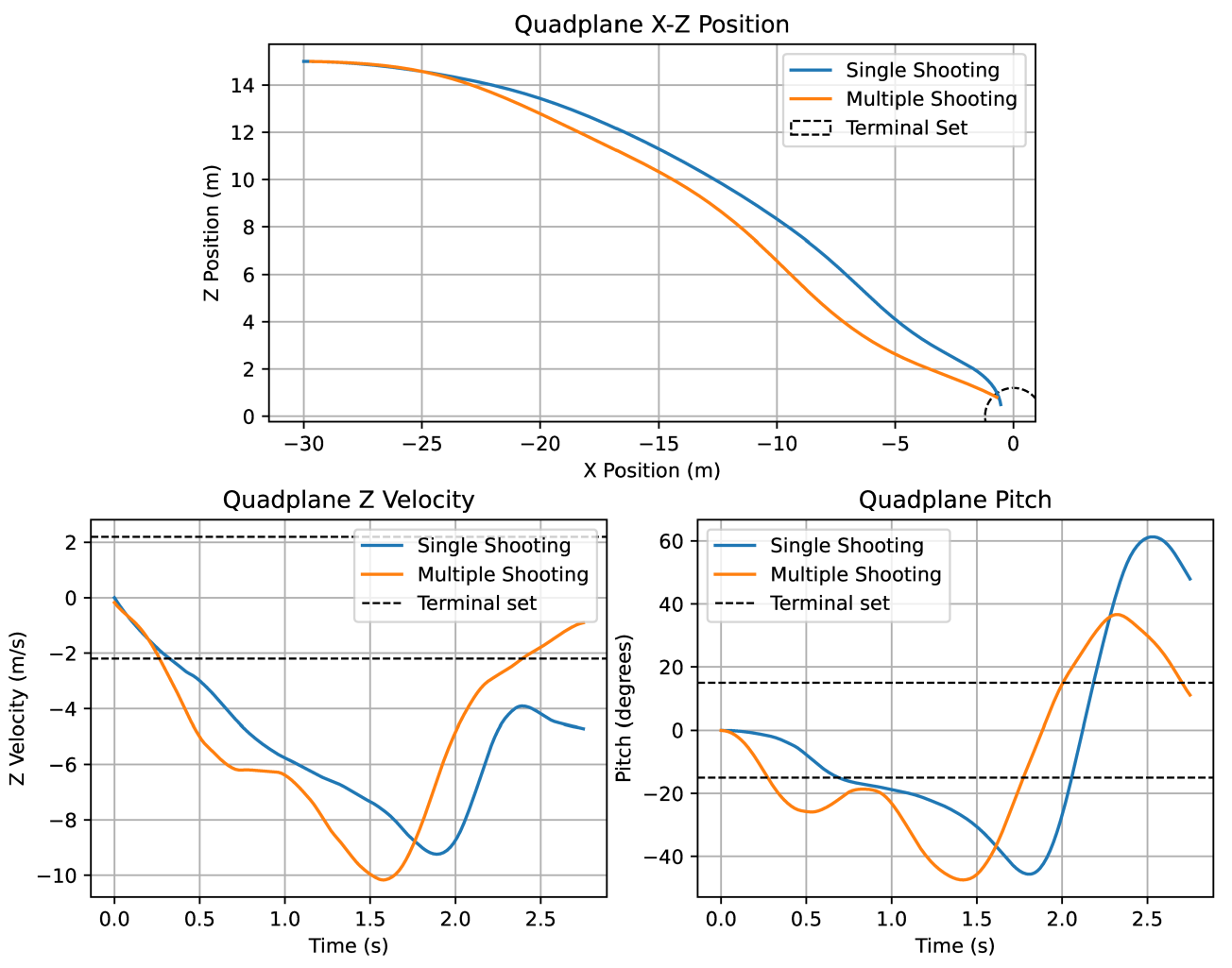}
    \caption{Quadplane Trajectories. Multiple shooting enables convergence into the desired terminal set.}
  \label{fig:quadplane-traj}
\end{figure}

We see in Table \ref{table:terminal} that convergence to the specified terminal set is only achievable via the multiple shooting method, and the multiple shooting method achieves much lower terminal costs on average. The trajectories are pictured in Fig. \ref{fig:quadplane-traj}. The cost versus rollouts is in Fig. \ref{fig:vtol-iters}.

\begin{table}[t]
\centering
\caption{Mean Terminal Cost at Convergence and Terminal Set Achieved}
\begin{tabular}{lcc}
\toprule
\textbf{Method} & \textbf{Cartpole Swingup} & \textbf{Quadplane Landing} \\
\midrule
MPPI                & 130.3 \xmark  & 115.2 \xmark \\
Single Shooting CEM & 123.4 \xmark & 53.3 \xmark \\
Multiple Shooting   & 23.8 \cmark & 20.3 \cmark \\
\bottomrule
\end{tabular}
\label{table:terminal}
\end{table}

 \section{DISCUSSION AND CONCLUSION}
In this work we proposed a novel stochastic multiple shooting algorithm that exhibits improved terminal set satisfaction compared to stochastic single shooting alternatives. The method enforces defects through local feedback policies which can be synthesized even for black-box dynamics, thus enabling application to model-based reinforcement learning.  The multiple shooting framework offers many areas for further investigation including hybrid dynamical systems, adaptive knot spacing, and various choices of optimizer and local feedback controller.

%%%%%%%%%%%%%%%%%%%%%%%%%%%%%%%%%%%%%%%%%%%%%%%%%%%%%%%%%%%%%%%%%%%%%%%%%%%%%%%

%%%%%%%%%%%%%%%%%%%%%%%%%%%%%%%%%%%%%%%%%%%%%%%%%%%%%%%%%%%%%%%%%%%%%%%%%%%%%%%%

%%%%%%%%%%%%%%%%%%%%%%%%%%%%%%%%%%%%%%%%%%%%%%%%%%%%%%%%%%%%%%%%%%%%%%%%%%%%%%%%

%\section*{ACKNOWLEDGMENT}

%%%%%%%%%%%%%%%%%%%%%%%%%%%%%%%%%%%%%%%%%%%%%%%%%%%%%%%%%%%%%%%%%%%%%%%%%%%%%%%%

%\section*{REFERENCES}

% BIBLIOGRAPHY
\bibliographystyle{IEEEtran}
\bibliography{references}

@article{polevoy2023probably,
  title={Probably Approximately Correct Nonlinear Model Predictive Control ({PAC-NMPC})},
  author={Polevoy, Adam and Kobilarov, Marin and Moore, Joseph},
  journal={IEEE Robotics and Automation Letters},
  volume={8},
  number={11},
  pages={7226--7233},
  year={2023},
  publisher={IEEE}
}

@article{mppi,
  author       = {Grady Williams and
                  Andrew Aldrich and
                  Evangelos A. Theodorou},
  title        = {Model Predictive Path Integral Control using Covariance Variable Importance
                  Sampling},
  journal      = {CoRR},
  volume       = {abs/1509.01149},
  year         = {2015},
  url          = {http://arxiv.org/abs/1509.01149},
  eprinttype   = {arXiv},
  eprint       = {1509.01149},
  bibsource    = {dblp computer science bibliography, https://dblp.org}
}

@ARTICLE{basecu2024swarm,
       author = {{Basescu}, Max and {Polevoy}, Adam and {Yeh}, Bryanna and {Scheuer}, Luca and {Sutton}, Erin and {Moore}, Joseph},
        title = "{Agile Fixed-Wing {UAV}s for Urban Swarm Operations}",
      journal = {IEEE Transactions on Field Robotics},
         year = 2024,
        month = jan,
       volume = {1},
        pages = {394-423},
          doi = {10.1109/TFR.2024.3496420},
       adsurl = {https://ui.adsabs.harvard.edu/abs/2024ITFR....1..394B}
}

@article{kobilarov2012cem,
  author       = {Kobilarov, Marin},
  title        = {Cross-entropy motion planning},
  journal      = {International Journal of Robotics Research},
  year         = 2012,
  volume       = 31,
  number       = 7,
  month        = jun,
  doi          = {10.1177/0278364912444543},
  url          = {https://doi.org/10.1177/0278364912444543},
}

@inproceedings{tracy2025trajectorybundle,
  title     = {The Trajectory Bundle Method: Unifying Sequential-Convex Programming and Sampling-Based Trajectory Optimization},
  author    = {Kevin Tracy and John Z. Zhang and Jon Arrizabalaga and Stefan Schaal and Yuval Tassa and Tom Erez and Zachary Manchester},
  booktitle = {Proceedings of the IEEE International Conference on Robotics and Automation (ICRA)},
  year      = {2026},
}

@misc{li2023unifiedperspectivemultipleshooting,
      title={A Unified Perspective on Multiple Shooting In Differential Dynamic Programming}, 
      author={He Li and Wenhao Yu and Tingnan Zhang and Patrick M. Wensing},
      year={2023},
      eprint={2309.07872},
      archivePrefix={arXiv},
      primaryClass={cs.RO},
      url={https://arxiv.org/abs/2309.07872}, 
}

@article{rao2010numericalmethods,
author = {Rao, Anil},
year = {2010},
month = {01},
pages = {},
title = {A Survey of Numerical Methods for Optimal Control},
volume = {135},
journal = {Advances in the Astronautical Sciences}
}

@article{bock1984,
title = {A Multiple Shooting Algorithm for Direct Solution of Optimal Control Problems*},
journal = {IFAC Proceedings Volumes},
volume = {17},
number = {2},
pages = {1603-1608},
year = {1984},
note = {9th IFAC World Congress: A Bridge Between Control Science and Technology, Budapest, Hungary, 2-6 July 1984},
issn = {1474-6670},
doi = {https://doi.org/10.1016/S1474-6670(17)61205-9},
author = {H.G. Bock and K.J. Plitt}
}

@article{ipopt,
author = {Wächter, Andreas and Biegler, Lorenz},
year = {2006},
month = {03},
pages = {25-57},
title = {On the Implementation of an Interior-Point Filter Line-Search Algorithm for Large-Scale Nonlinear Programming},
volume = {106},
journal = {Mathematical programming},
doi = {10.1007/s10107-004-0559-y}
}

@misc{manchester2017controlcontractionmetricsconvex,
      title={Control Contraction Metrics: Convex and Intrinsic Criteria for Nonlinear Feedback Design}, 
      author={Ian R. Manchester and Jean-Jacques E. Slotine},
      year={2017},
      eprint={1503.03144},
      archivePrefix={arXiv},
      primaryClass={cs.SY},
      url={https://arxiv.org/abs/1503.03144}, 
}

@misc{yang2019dataefficientreinforcementlearning,
      title={Data Efficient Reinforcement Learning for Legged Robots}, 
      author={Yuxiang Yang and Ken Caluwaerts and Atil Iscen and Tingnan Zhang and Jie Tan and Vikas Sindhwani},
      year={2019},
      eprint={1907.03613},
      archivePrefix={arXiv},
      primaryClass={cs.LG},
      url={https://arxiv.org/abs/1907.03613}, 
}

@ARTICLE{robustmppi,
  author={Gandhi, Manan S. and Vlahov, Bogdan and Gibson, Jason and Williams, Grady and Theodorou, Evangelos A.},
  journal={IEEE Robotics and Automation Letters}, 
  title={Robust Model Predictive Path Integral Control: Analysis and Performance Guarantees}, 
  year={2021},
  volume={6},
  number={2},
  pages={1423-1430},
  doi={10.1109/LRA.2021.3057563}}

@misc{wang2021variationalinferencempcusing,
      title={Variational Inference {MPC} using {Tsallis} Divergence}, 
      author={Ziyi Wang and Oswin So and Jason Gibson and Bogdan Vlahov and Manan S. Gandhi and Guan-Horng Liu and Evangelos A. Theodorou},
      year={2021},
      eprint={2104.00241},
      archivePrefix={arXiv},
      primaryClass={cs.LG},
      url={https://arxiv.org/abs/2104.00241}, 
}

@misc{yi2024covompc,
      title={{CoVO-MPC}: Theoretical Analysis of Sampling-based {MPC} and Optimal Covariance Design}, 
      author={Zeji Yi and Chaoyi Pan and Guanqi He and Guannan Qu and Guanya Shi},
      year={2024},
      eprint={2401.07369},
      archivePrefix={arXiv},
      primaryClass={cs.LG}
}

@misc{pan2024modelbaseddiffusiontrajectoryoptimization,
      title={Model-Based Diffusion for Trajectory Optimization}, 
      author={Chaoyi Pan and Zeji Yi and Guanya Shi and Guannan Qu},
      year={2024},
      eprint={2407.01573},
      archivePrefix={arXiv},
      primaryClass={cs.RO},
      url={https://arxiv.org/abs/2407.01573}, 
}

@article{dircol,
author = {HARGRAVES, C. and Paris, Stephen},
year = {1987},
month = {07},
pages = {338-342},
title = {Direct Trajectory Optimization Using Nonlinear Programming and Collocation},
volume = {10},
journal = {AIAA J. Guidance},
doi = {10.2514/3.20223}
}

@misc{yin2022riskawaremodelpredictivepath,
      title={Risk-Aware Model Predictive Path Integral Control Using Conditional Value-at-Risk}, 
      author={Ji Yin and Zhiyuan Zhang and Panagiotis Tsiotras},
      year={2022},
      eprint={2209.12842},
      archivePrefix={arXiv},
      primaryClass={cs.RO},
}

@article{Mayne1966ASG,
  title={A Second-order Gradient Method for Determining Optimal Trajectories of Non-linear Discrete-time Systems},
  author={David Q. Mayne},
  journal={International Journal of Control},
  year={1966},
  volume={3},
  pages={85-95},
}

@conference{Howell-2019-122091,
author = {Taylor Howell And Brian Jackson And Zac Manchester},
title = {{ALTRO}: A Fast Solver for Constrained Trajectory Optimization},
booktitle = {Proceedings of (IROS) IEEE/RSJ International Conference on Intelligent Robots and Systems},
year = {2019},
month = {November},
pages = {7674 - 7679},
}

@article{PELLEGRINI2020686,
title = {A multiple-shooting differential dynamic programming algorithm. Part 1: Theory},
journal = {Acta Astronautica},
volume = {170},
pages = {686-700},
year = {2020},
issn = {0094-5765},
doi = {https://doi.org/10.1016/j.actaastro.2019.12.037},
url = {https://www.sciencedirect.com/science/article/pii/S0094576519314705},
author = {Etienne Pellegrini and Ryan P. Russell}
}

@article{Bordalba_2023,
   title={Direct Collocation Methods for Trajectory Optimization in Constrained Robotic Systems},
   volume={39},
   ISSN={1941-0468},
   url={http://dx.doi.org/10.1109/TRO.2022.3193776},
   DOI={10.1109/tro.2022.3193776},
   number={1},
   journal={IEEE Transactions on Robotics},
   publisher={Institute of Electrical and Electronics Engineers (IEEE)},
   author={Bordalba, Ricard and Schoels, Tobias and Ros, Lluis and Porta, Josep M. and Diehl, Moritz},
   year={2023},
   month=Feb, pages={183–202} }

@article{betts1998survey,
  title={Survey of numerical methods for trajectory optimization},
  author={Betts, John T},
  journal={Journal of guidance, control, and dynamics},
  volume={21},
  number={2},
  pages={193--207},
  year={1998}
}

@article{pardo2016evaluating,
  title={Evaluating direct transcription and nonlinear optimization methods for robot motion planning},
  author={Pardo, Diego and M{\"o}ller, Lukas and Neunert, Michael and Winkler, Alexander W and Buchli, Jonas},
  journal={IEEE Robotics and Automation Letters},
  volume={1},
  number={2},
  pages={946--953},
  year={2016},
  publisher={IEEE}
}

@incollection{kraft1985converting,
  title={On converting optimal control problems into nonlinear programming problems},
  author={Kraft, Dieter},
  booktitle={Computational mathematical programming},
  pages={261--280},
  year={1985},
  publisher={Springer}
}

@article{bock1984multiple,
  title={A multiple shooting algorithm for direct solution of optimal control problems},
  author={Bock, Hans Georg and Plitt, Karl-Josef},
  journal={IFAC Proceedings Volumes},
  volume={17},
  number={2},
  pages={1603--1608},
  year={1984},
  publisher={Elsevier}
}

@book{bryson2018applied,
  title={Applied optimal control: optimization, estimation and control},
  author={Bryson, Arthur Earl},
  year={2018},
  publisher={Routledge}
}

@article{ozaki2020tube,
  title={Tube stochastic optimal control for nonlinear constrained trajectory optimization problems},
  author={Ozaki, Naoya and Campagnola, Stefano and Funase, Ryu},
  journal={Journal of Guidance, Control, and Dynamics},
  volume={43},
  number={4},
  pages={645--655},
  year={2020},
  publisher={American Institute of Aeronautics and Astronautics}
}

@inproceedings{todorov2005generalized,
  title={A generalized iterative {LQG} method for locally-optimal feedback control of constrained nonlinear stochastic systems},
  author={Todorov, Emanuel and Li, Weiwei},
  booktitle={Proceedings of the 2005, American control conference, 2005.},
  pages={300--306},
  year={2005},
  organization={IEEE}
}

@article{howell2021direct,
  title={Direct policy optimization using deterministic sampling and collocation},
  author={Howell, Taylor A and Fu, Chunjiang and Manchester, Zachary},
  journal={IEEE Robotics and Automation Letters},
  volume={6},
  number={3},
  pages={5324--5331},
  year={2021},
  publisher={IEEE}
}

@article{scs,
    author       = {Brendan O'Donoghue and Eric Chu and Neal Parikh and Stephen Boyd},
    title        = {Conic Optimization via Operator Splitting and Homogeneous Self-Dual Embedding},
    journal      = {Journal of Optimization Theory and Applications},
    month        = {June},
    year         = {2016},
    volume       = {169},
    number       = {3},
    pages        = {1042-1068},
    url          = {http://stanford.edu/~boyd/papers/scs.html},
}

@article{andersen2013cvxopt,
  title={{CVXOPT}: Python software for convex optimization},
  author={Andersen, Martin S and Dahl, Joachim and Vandenberghe, Lieven and others},
  journal={URL https://cvxopt. org},
  volume={64},
  year={2013}
}

\end{document}